\documentclass[conference]{IEEEtran}
\IEEEoverridecommandlockouts
\usepackage{perpage} 
\MakePerPage{footnote} 
\usepackage{cite}
\usepackage{amsmath,amssymb,amsfonts}
\usepackage{algorithmic}
\usepackage{graphicx}
\usepackage{textcomp}
\usepackage{xcolor}
\def\BibTeX{{\rm B\kern-.05em{\sc i\kern-.025em b}\kern-.08em
    T\kern-.1667em\lower.7ex\hbox{E}\kern-.125emX}}
\usepackage{multicol}
\usepackage{blindtext}
\usepackage{tabularx}
\usepackage{tabularray}

\begin{document}
\onecolumn
\textbf{IEEE Copyright Notice}

© 2023 IEEE.  Personal use of this material is permitted.  Permission from IEEE must be obtained for all other uses, in any current or future media, including reprinting/republishing this material for advertising or promotional purposes, creating new collective works, for resale or redistribution to servers or lists, or reuse of any copyrighted component of this work in other works.
\twocolumn
\newpage

\title{Temporal Link Prediction Using Graph Embedding Dynamics}

\author{\IEEEauthorblockN{1\textsuperscript{st} Sanaz Hasanzadeh Fard}
\IEEEauthorblockA{\textit{dept. Computer Science and Engineering} \\
\textit{Michigan State University}\\
East Lansing, Michigan \\
hasanzad@msu.edu}
\and
\IEEEauthorblockN{2\textsuperscript{nd} Mohammad Ghassemi}
\IEEEauthorblockA{\textit{dept. Computer Science and Engineering} \\
\textit{Michigan State University}\\
East Lansing, Michigan \\
ghassem3@msu.edu}
}

\maketitle

\begin{abstract}
  Graphs are a powerful representation tool in machine learning applications, with link prediction being a key task in graph learning. Temporal link prediction in dynamic networks is of particular interest due to its potential for solving complex scientific and real-world problems. Traditional approaches to temporal link prediction have focused on finding the aggregation of dynamics of the network as a unified output. In this study, we propose a novel perspective on temporal link prediction by defining nodes as Newtonian objects and incorporating the concept of velocity to predict network dynamics. By computing more specific dynamics of each node, rather than overall dynamics, we improve both accuracy and explainability in predicting future connections. We demonstrate the effectiveness of our approach using two datasets, including 17 years of co-authorship data from \textit{PubMed}. Experimental results show that our temporal graph embedding dynamics approach improves downstream classification models' ability to predict future collaboration efficacy in co-authorship networks by 17.34\% (AUROC improvement relative to the baseline model). Furthermore, our approach offers an interpretable layer over traditional approaches to address the temporal link prediction problem.
\end{abstract}

\begin{IEEEkeywords}
link prediction, temporal graph embedding, scientific collaborations network, co-authorship prediction, network dynamics
\end{IEEEkeywords}

\section{Introduction}
Several recent works in machine learning research have demonstrated the impressive utility of graphs to flexibly represent complex phenomenon \cite{b1, kazemi2020representation} including biological networks \cite{sulaimany2018link}, brain networks \cite{25}, climate networks \cite{30,31,32}, multi-agent systems \cite{hasanzadeh2019two}, computer networks \cite{33}, and social networks \cite{27,34,35}. One common graph inference problem is that of link prediction \cite{12} - given a pair of vertices, we would like to determine the probability of an edge's existence now, or at some point in the future. This problem is in the intersection of graph theory, complex network analysis \cite{7}, and the network science domain \cite{8,9,10,11}. Traditional link prediction approaches focus on static link prediction \cite{kumar2020link} while the mentioned networks have an evolving nature and static link prediction techniques cannot be applied to complex networks with time-varying dynamics \cite{kazemi2020representation}. There has been a significant interest among researchers in addressing the problem of temporal link prediction \cite{kazemi2020representation, divakaran2020temporal} in more recent years. In the following, we address the problem of link prediction in dynamic networks in more detail and the proposed techniques for that in the literature.

In the study of dynamic graphs and temporal link prediction, the dimension of time is added to the problem of static link prediction and has changed the nature of this problem. Before addressing the temporal link prediction techniques, we need to know how different approaches model network changes over time. There are two data models to describe time-varying dynamics: discrete-time models and continuous-time \cite{xue2022dynamic}. Discrete-time models for temporal link prediction consider time as a series of discrete intervals, such as days or weeks, and predict the probability of a link forming between two nodes within each interval. Continuous-time models, on the other hand, consider time as a continuous variable and predict the probability of a link forming at any given point in time. Continuous-time models are often more flexible and accurate but can be more computationally intensive than discrete-time models.

There are different techniques in the literature for the problem of temporal link prediction. One general direction of techniques for temporal link prediction is temporal graph embedding techniques \cite{barros2021survey, wang2022survey}; these techniques aim to learn low-dimensional representations of nodes in a temporal graph that capture their temporal dynamics. Temporal graph embedding techniques include methods such as Temporal Random Walk \cite{starnini2012random}, Dynamic Triad Embedding \cite{zhou2018dynamic}, and Temporal Graph Convolutional Networks
\cite{zhao2019t}. The next series of approaches for the problem of temporal link prediction is Recurrent Neural Networks (RNNs); RNNs are particularly useful when the temporal dynamics of the graph are sequential in nature, such as in time-series data \cite{nasiri2022mfrfnn}. In this approach, each node is represented as a sequence of vectors over time, where each vector captures the node's state at a particular time-step. The RNN is trained to predict the next state of each node based on its previous states and the states of its neighboring nodes. RNN-based approaches for temporal link prediction include methods such as RNNs for Link Prediction (RNNLP) \cite{hajiramezanali2019variational} and Temporal Attention-based LSTM (TALSTM) \cite{li2020hierarchical}. 

Matrix Factorization (MF) method \cite{ma2018graph} is another temporal link prediction technique. In MF, the adjacency matrix of the graph is factorized into two low-rank matrices representing the latent features of the nodes and the latent features of the time periods. The latent features of the nodes capture their underlying properties, while the latent features of the time periods capture the temporal dynamics of the graph. MF can be used to predict future links by estimating the missing entries in the adjacency matrix. MF-based approaches for temporal link prediction include methods such as Temporal Regularized MF (TRMF) \cite{yu2016temporal} and Dynamic MF (DMF) \cite{ma2018graph}. We discuss recently proposed temporal link prediction approaches in the experiments section in order to compare our approach against them.

Our particular interest in this work is temporal link prediction in scientific co-authorship networks \cite{chuan2018link}. The ability to predict \textit{prospective} collaborative structures in scientific communities may enable the formation of more productive scientific networks \cite{1}, generating higher quality research output and publications \cite{2}. Researchers can be productive as individuals, while the outcome of their teamwork can exceed the sum of their individual efforts \cite{2,6}. Our main contribution in this work is proposing a simple yet efficient explainable layer over graph embedding techniques to improve temporal link prediction performance; this work consists of 1- defining each node as an object in Euclidean space, 2- calculating the temporal velocity sequence of nodes (i.e., objects), 3- predicting future velocity of nodes, 4- computing future location of nodes using location-velocity formula, and finally, 5- predicting future connections between nodes base on their predicted placement. The rest of this paper is organized as follows: the second section elaborates on the proposed methodology, the third section describes designed experiments, the fourth section brings the results, the next section is assigned to the discussion, and the last section is about the conclusion and future directions of our research.

\section{Method}
Our goal is to predict future scientific collaborations based on previous years' co-authorship networks. In this section, we provide more details on the proposed approach to accomplish this goal. 

\subsection{Illustrative Overview}
Fig. 1 (top) illustrates the prediction process of the future velocity, followed by (middle) the prediction of the authors' location, and finally (bottom) link prediction, using their predicted future locations. Fig. 1 (top) shows the velocity and direction of node movements; the location of embedded nodes (feature vector) in the latent space, for two consecutive years, is shown. Each color represents one author (each author's color stays the same during all years); circles and squares represent authors' locations in the first and second years respectively. Temporal embedding transfers the graph of each year to the same latent space; so, after transferring the second year, the movement direction of each embedded node in the latent space can be obtained. Fig. 1 (top) also shows the movement direction of seven authors in the latent space; the arrow represents the velocity vector of each node moving

\begin{figure}
  \makeatletter
  \def
  \@captype{figure}
  \includegraphics[width=0.5\textwidth]{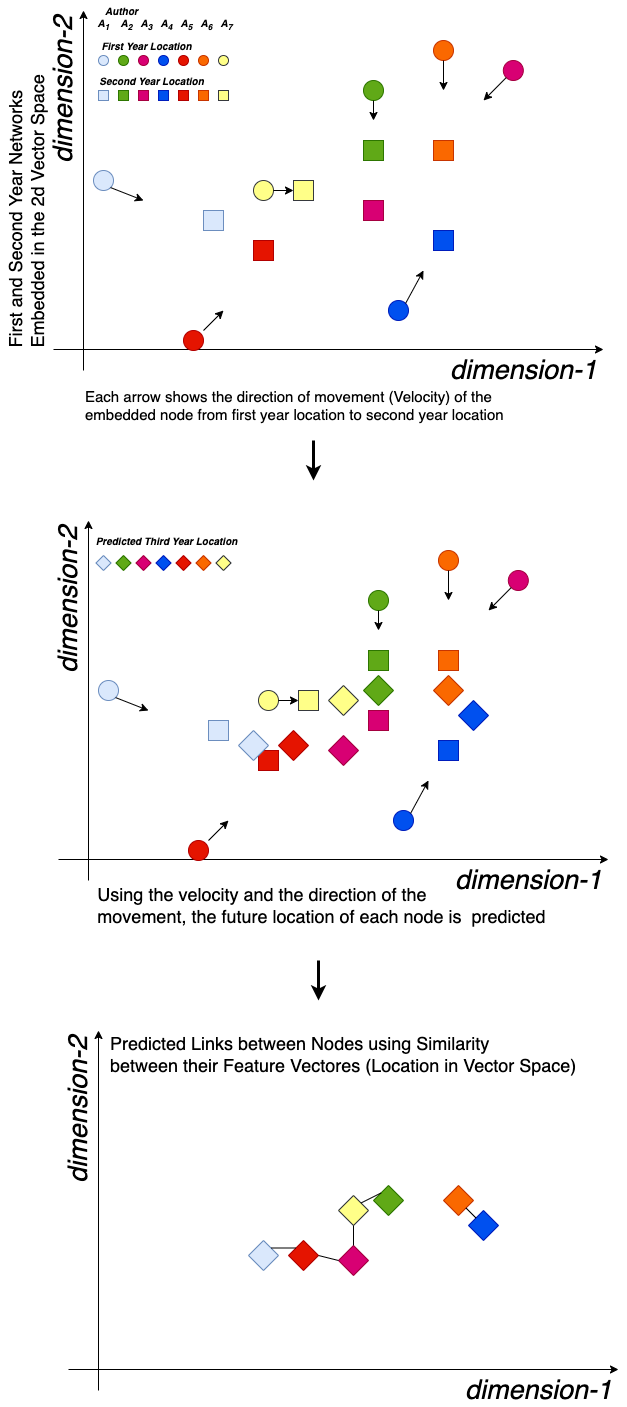}
  \makeatother
  \caption{Velocity prediction, followed by link prediction. (Top) Tracking the movement of nodes in the embedding space from their location in the first year moving to the second year. ( Middle) Predicting the future location of nodes in the embedding space based on their current location (second/last year, shown by squares) and the movement's dynamics; i.e., velocity. (Bottom) Predicting links between the predicted location of nodes in the future based on their similarity in the embedding space.}
  \label{fig:method}
\end{figure}

\newpage
 from its location in the first year (circles) to its location in the second year (squares).

Having the current and previous locations of nodes and knowing the time interval of the movement, we calculate the velocity of each author's movement in the latent space. 
In Fig. 1 (middle), diamonds represent the predicted location of authors using their velocity and current location. These predicted locations are on the curve of the trajectory of node movement across time (as estimated by LSTM), and relative to their current and previous locations. Fig. 1 (bottom) shows predicted links using the predicted location of authors and the similarity between these predicted locations (similarity and closeness between their predicted feature vectors).

\subsection{Graph Embedding Approach}
To represent the authorship networks as a graph, each author and each collaboration between author pairs are represented as nodes and links respectively. More specifically, for each given year, a given pair of authors have been deemed "collaborators" if their names jointly appeared on one or more papers published during that year. Given our node and link definitions, we generated one co-authorship network for each calendar year. Our goal was to develop a model to predict the collaboration structure (i.e., the links) in a given year, based on the collaboration structure of multiple years prior.

Following the representation of our data as a set of annual co-authorship graphs, we performed graph embedding. The role of the embedding stage was to represent each author, in each year, as a point in a vector space. Graph embedding was performed using the DynamicGem package \cite{goyal2018graph}; more specifically, the embeddings were generated using the Dynamic AERNN method \cite{21}. This method is a common approach for temporal embedding that enables both compressions (via a fully connected encoder) and representation of the temporal dynamics (via LSTMs) \cite{goyal2018graph}. Importantly, the recurrent structure of AERNN allows it to account for embedding dependencies \textit{across} years. That is, the network of the second year is transferred to the latent space with respect to the nodes’ placement in the previous year. This is important to trace the networks' dynamics across time.

\subsection{Essential Notation}
Before elaborating on the proposed approach, we define notation: given graph $G(V, E)$, where $V$ is the set of nodes and $E$ is the set of links between nodes, and having states of the graph in $0 \rightarrow t$ consecutive \textit{time-steps}, our goal is to predict the state of the graph at time-step $t+1$. Set $S = \{s_0, s_1, ..., s_t\}$ represents the set of $t+1$ consecutive states of the graph, where each state $s_i$ represents the graph at time-step $i$. Using AERNN we represent each node $p \in V$, at $s_i \in S$, in a latent space as a vector $r_p^i$ at each time-step $i$, by learning a function $f: p^i \rightarrow{ r_p^i}$, which maps each node from graph-structured data to a vector in \textit{Euclidean} space, $\mathbb{R}^d$ where $d$ is the dimension of latent (i.e., embedding) space. 

\subsection{Proposed Innovation}
The novelty of our work is an extension of \textit{Temporal Embedding Approaches} that considers the embedded nodes as \textit{Newtonian Objects} characterized by their position, as well as their $Velocity$ in the embedding space. That is, we use the difference in embedding position over multiple years to infer the dynamics of the network, and use these dynamics to predict the future location (and hence, the connectivity) of the nodes. The velocity formula of node $p$, moving from time-step $i-1$ to $i$ is represented in (1). 
\begin{equation}
    \forall p | p\in{V} \text{ and } \forall (s_{i-1},s_i) | s_{i-1}, s_i \in S:
    $$
    $$
      Vel_p^{i-1 \rightarrow{i}} = r_p^{i}-r_p^{i-1}
\end{equation}

One can predict a node's velocity moving toward the next state by feeding a sequence of consecutive velocities of the node, in time-series form, to a time-series prediction model; in our case, the velocity prediction model used an LSTM. Our velocity predictor model is a 3-layer recurrent network with $512, 256$ LSTM units in the first and second layers, respectively; the number of LSTM units in the third layer changes according to embedding size. After embedding all nodes across all years, we have $\forall p \in{V}: r_p = \{r_p^0, r_p^1, r_p^2, ..., r_p^t\}$. By passing this sequence to (1) for each node, the velocity sequence for each node is generated as $ \forall p \in{V}: Vel_p^{1..t} = \{Vel_p^1, Vel_p^2, ..., Vel_p^t\}$. Note that the length of $Vel_p^{1..t}$ is $t$ which is 1 unit shorter than $r_p$. Here, we have the velocity of node $p$ when it arrives at $s_1$ as $Vel_p^1$ (moving from $s_0$ to $s_1$). $Vel_p^t$ is velocity of $p$ at $s_t$ and goal is predicting $p$'s velocity moving towards $s_{t+1}$ in the future. 

After predicting the dynamics of each node (i.e., velocity) we did a 2-dimensional aggregation for each node: first, aggregation over time-steps (vertical) and second, neighbor aggregation in the last time-step (horizontal). Vertical aggregation captures historical-temporal information of the node's dynamics using a weighted aggregation function over the past $h$ velocities of the node, including the predicted velocity; this aggregated velocity represents the temporal information of the dynamics. The aggregation formula is shown in (2); the equation performs a weighted sum of the past $h$ node velocities, biased to favor more recent velocities.

The goal of velocity aggregation is to find the direction of node movements; we compute a weighted resultant of the node's last $h$ velocities to keep the final velocity on the curve of node movements. Equation (2) captures the temporal order of velocities by weighting them. This way, recent dynamics, which are more deterministic, get higher weights.
\begin{equation}
    Vel_p^{Agg_{t+1}} = \frac{2}{h \times (h+1)} \Sigma_{\eta = 1, \tau = t-h}^{\eta = h, \tau = t+1} {\eta \times Vel_p^{\tau}}
\end{equation}
Using the aggregated velocity, we initialize the location of nodes at time-step $t+1$ as we can see in (3). In this equation, \textit{Init} indicates the initialization value.

\begin{equation}
    Loc_p^{Init_{t+1}} = Vel_p^{Agg_{t+1}} + r_p^t
\end{equation}

The location of each node is 1- dependent on its history which we addressed by vertical aggregation, and 2- dependent on interactions between nodes with the rest of the network; a given node should be closer to its neighbors and far from nodes that are not in its one-hop neighbors\footnote{One-hop neighbors of each node are its directly connected neighbors.}. Horizontal aggregation captures interactions between each node and the rest of the network. Zhang et al. \cite{zhang2018link} proved that the local network of a node provides enough information on node dynamics in order to predict its behavior. According to Zhang et al. \cite{zhang2018link}, a second aggregation over $Loc^{Init}$ of one-hop neighbors of the node, can be used to capture the interaction of the node's local network. In our case, the aggregation function is the average over $Loc^{Init}$ of the node's one-hop neighbors and the node itself as shown in (4).

\begin{equation}
    Loc_p^{Agg_{t+1}} = \frac{1}{|N|+1}\Sigma_{u \in N(p) \cup p} {Loc_u^{Init_{t+1}}}
\end{equation}
where $N$ is the set of one-hop neighbors of node $p$.

The last update to the nodes is done by horizontal aggregation and provides us with final locations; link prediction is done based on these locations. It is done by measuring the level of similarity of each pair of nodes.

After predicting the location of nodes, \textit{Euclidean Similarity} is applied to each pair of (location of) nodes in order to determine the level of similarity between them; this similarity level will be proportional to the probability that the nodes will contain a link in the future and can be used for edge classification \footnote{Following re-scaling between 0 and 1}. In this research, we have used the \textit{Area Under the Receiver Operator Curve (AUROC)} and the \textit{Area Under the Precision-Recall Curve (AUPRC)} as performance metrics. 

\section{Experiments}
Experiments were designed with three goals; the primary goal of the first series of experiments was to compare the performance of the proposed approach against AERNN as the baseline. The second series of experiments investigated the effect of model hyper-parameters on the performance of both approaches. The last series of experiments examines the performance of the proposed approach on a common dataset in the field of temporal link prediction against the recently proposed approaches for this problem.

\subsection{Dataset}
Data for the first and second series of experiments was sourced from BRAINWORKS, a curated subset of 40 Million publicly available scientific papers, and their accompanying meta-data (authors, citations, funding sources, etc.) \cite{Ghassemi2021BRAINWORKS}. We extracted all publications available in BRAINWORKS from 2005-2021; this provided a total of 1,733,813 papers. From each paper, we extracted the set of all authors; in total, there were 6,305,889 authors over the 17-year period, an average of 370,934 per year. For related experiments, we focused our analysis on a subset of authors that had published at least one paper per year (random 3,000 out of 13,655 authors), over the 17-year period spanned by our data; we emphasize here that this inclusion criterion is based on the common idea of considering known nodes when studying link prediction problem \cite{singer2019node, sankar2020dysat}. 


\subsection{Performance Metrics}
We have used \textit{Area Under the Receiver Operator Curve (AUROC)} and \textit{Area Under the Precision-Recall Curve (AUPRC)} as performance metrics. AUROC and AUPRC were evaluated on multiple random samples (n=3,000 nodes) drawn from the 13,655 authors in our data.
More specifically, we report the results of experiments over $10$ randomly selected samples of $3,000$ authors; all authors published at least one paper in the 17-year time-frame. We evaluated the ability of the normalized ($[0,1]$) Euclidean similarity measures generated by our model to predict co-authorship structure in 2021, given data from 2005-2020.

\subsection{Model Training and Validation Approach}
Train and test sets were generated as follows: from each random sample of authors, the induced subgraph was obtained (across all time) and embedded using the AERNN encoder with lookback = 1 (lookback is set to 1 to generate embeddings for our approach across all experiments, any other referral to lookback is about its setting for AERNN in the experiments); since social networks are large-scale sparse graphs, the number of connected pairs of nodes (positive samples) is notably smaller than the number of unconnected nodes (negative samples). Hence, we included all positive samples and randomly selected the same number of negative samples, per year. Positive and negative samples were labeled +1 and -1, respectively. Data from the years 2005-2020 were used as training data, while data from 2021 was used for testing.
\subsection{Comparison of Proposed Approach Against the Baseline}

To compare the proposed approach against the baseline, we fed graphs from the years 2015-2020 as training data to both methods. So, the length of the series was set to 6. We set the number of previous graphs for learning to $3$ for the baseline (lookback = 3). The history length for velocity aggregation for the proposed approach was set to 3. The size of the embedding dimension was set to $128$ for both methods. 
In order to investigate whether the proposed approach improves the quality of the embeddings generated by the baseline, we generated the final embedding of the baseline (appeared in figures as \textit{Raw Embedding} which is the output of the encoder of AERNN and the input to our framework) and our approach. \textit{Euclidean Similarity} function was applied to the Raw Embedding and locations (i.e., embeddings) predicted by our proposed method to generate connection probabilities. We compared these probabilities against predicted links by AERNN. 

\subsection{Hyper-Parameters Analysis}
In this research, we had three main hyper-parameters. The \textit{embedding size (dim)}, which is the primary parameter of approaches that have an embedding step as part of their framework, \textit{time-series length (l)}, the parameter determining the length of needed previous states of the network in order to learn long-term dependencies, and \textit{history (h)}, the parameter controlling the length of recent history directly involved in updating dynamics (dominant history of dynamics). 

The first evaluated parameter is the size of the embedding space. The goal is to find out how changing the dimension of the embedding space affects the performance and whether the approach is robust to changes in dimension; also, we evaluated whether we needed a higher dimension size to reach higher performance. As the primary setting stands for $l = 6$, $lookback = h = 3$, we did experiments for $dim = \{32, 64, 128, 256\}$.

The next evaluated parameter is $l$, the length of time-series. Analyzing $l$ leads to finding the optimum length for learning long-term dynamics. Shorter series results in time efficiency as well as eliminating data redundancy, in case the performance of shorter series is still high; if a smaller number of time-steps is enough for learning, we do not need extra data and could do experiments more efficiently, in terms of both time and computation. Also, data shortage is a common issue in machine learning problems; so, if a shorter sequence could result in sufficient performance, we can solve a greater range of problems having smaller data. We did experiments by setting $lookback = h = 3$ and $dim = 128$. We tested $6, 9, 12$, and $15$ for $l$.

The last evaluated parameter is $h$; history length. The set of studied values for $h$ is $\{2, 3, 4, 5, 6, 7, 8, 9\}$; the goal of this experiment is to study the length of recent temporal dependencies, that are directly affecting future behavior. Finding the point where dynamic dependencies vanish is another goal. $l$ is set to $12$ and $dim$ to $128$. With this range, we can compare small history lengths against higher ones. 

\subsection{Comparison of the Proposed Approach with State-of-the-art Methods}

In this section, we expanded our experiments to evaluate the performance of our approach against a wider range of techniques. 
We used \textit{Hyper} dataset \cite{isella2011s,SocioPatterns} for this section. Hyper (Hypertext2009) is a network of face-to-face contact of the participants of the ACM Hypertext Conference 2009. Link prediction techniques that were used for comparison include GraphSAGE \cite{hamilton2017inductive}, CTDNE \cite{nguyen2018continuous}, SGNN \cite{ma2020streaming}, TREND \cite{wen2022trend}, and DMAB \cite{wu2023temporal}. GraphSAGE is a graph representation learning algorithm that learns node embeddings by aggregating information from its neighboring nodes in a graph. The GraphSAGE algorithm consists of two main components: 1. Aggregation function: This function aggregates the feature vectors of the neighboring nodes of a given node to generate a summary vector that represents the local neighborhood of the node. 2. Update function: This function takes the summary vector generated by the aggregation function and updates the embedding of the node.

CTDNE (Continuous-Time Dynamic Network Embedding) is a graph representation learning algorithm that learns node embeddings in dynamic graphs, where the graph structure changes over time. CTDNE models the evolution of the graph using continuous-time Markov processes and learns embeddings that capture both the static and temporal aspects of the graph. The CTDNE algorithm consists of three main components: 1. Transition probability function: This function models the probability of a node transitioning to a new state at any given time. 2. Objective function: This function measures the quality of the learned embeddings by comparing them to the observed transitions in the graph. 3. Optimization algorithm: This algorithm updates the embeddings to minimize the objective function.

SGNN (i.e., Streaming Graph Neural Networks) regards temporal networks as streaming data and uses a message-passing mechanism for representation learning, which updates node information by capturing
the sequential information of edges (interactions), time intervals
between edges, and information propagation. TREND (i.e., Temporal event and node dynamics for graph representation learning) is a GCN-based method inspired by the self-exciting effect of the Hawkes process that captures the individual and collective characteristics of events by integrating event and node popularity, driving more precise modeling of the dynamic process.

DMAB: In this approach, to explain network evolution more interpretably, two dynamic properties of nodes are extracted and quantified: activity and loyalty. Activity is the basic ability of a node to obtain links, and loyalty is its ability to maintain its current link state. Based on the activity and loyalty properties of nodes, the Develop-Maintain Activity Backbone (DMAB) model performs link prediction. The DMAB model integrates these two modules for link prediction. The Activity Backbone describes the inclination of nodes to create links based on their activity level. The Maintain-Develop Module (DMM) describes whether a node tends to develop new friends or maintain old relations based on their loyalty level.

\subsubsection{Experimental Setting}
For all the approaches, we divided links based on the time they appeared in ascending order with the ratio 3:1 to train and test sets. The maximum number of {1, 2, 3, 4, 5}-hop neighbor nodes for GraphSAGE is {25, 10, 10, 10, 10}. For CTDNE, temporal neighbor selection has been set to unbiased distribution \cite{nguyen2018continuous}. For SGNN, the maximum propagation size is set to 50. For all the baseline methods, their best performance across Average, Cosine, Hadamard, Weighted-L1, and Weighted-L2 similarity functions is reported. For our approach, we set history = 3 and divided the training set into 7 overlapping timesteps based on validation on set \{6, 7, 8, 9\}. The embedding size is 128 for all methods.

\section{Results}
In this section, we will first bring the result of the comparison of the proposed approach against the baseline, i.e., AERNN. Then, the result of changing \textit{dim}, $l$, and $h$ over experiments will be shown. Table $1$ represents the result of link prediction experiments on Raw Embeddings, AERNN, and the proposed approach. Based on table 1, our proposed approach has outperformed the baseline in terms of both AUROC and AUPRC scores. Also, based on experimental results, our approach highly improved the baseline's Raw Embedding quality. Table 1 indicates that the proposed approach has improved the AUROC score of the baseline by $17.34\%$ (0.119 absolute increase). 

\begin{table}
\centering
\caption{COMPARISON OF PROPOSED APPROACH AGAINST AERNN}
\begin{tblr}{
  width = \linewidth,
  colspec = {Q[469]Q[212]Q[206]},
  column{2} = {c},
  column{3} = {c},
  hline{1-2,5} = {-}{},
}
\textbf{Approach} & \textbf{AUROC}  & \textbf{AUPRC}  \\
AERNN             & 0.687           & 0.7911          \\
Raw Embedding     & 0.3475~         & 0.3974~         \\
Proposed Approach & \textbf{0.8062} & \textbf{0.8462} 
\end{tblr}
\end{table}

Figs. 2 and 3 show how changing the embedding dimension affects the performance of the two approaches, as well as the performance of Raw Embedding, in terms of AUROC and AUPRC scores, respectively. For this section, we fix the setting for $h$ and $l$ to $3$ and $6$ respectively, and repeated the experiment with setting $dim$ to $32$, $64$, $128$, and $256$. Fig. 2 represents that the best results for Raw Embedding, AERNN, and the proposed approach have been achieved by setting $dim = 32$, $dim = 256$, and $128$, respectively. These results are true for both AUROC and AUPRC scores. Figs. 4 and 5 demonstrate the experimental results of the effect of changing the length of time-series on link prediction performance. Setting is as follows: $dim = 128$, $h = 3$, and $l \in \{6, 9, 12, 15\}$. 

\begin{figure}[t!]
  \centering
  \includegraphics[width=1\linewidth]{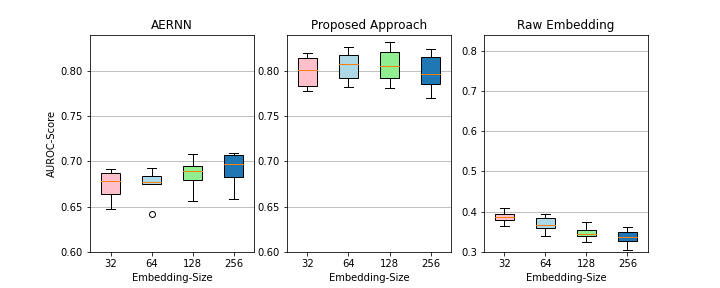}
  \caption{Performance Evaluation of Different Embedding Sizes in terms of AUROC Score }
  \label{fig:EmbSizeAUPRC}
\end{figure}

\begin{figure}[t!]
  \centering
  \includegraphics[width=1\linewidth]{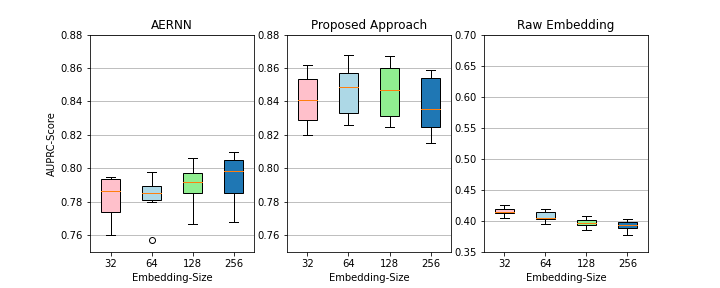}
  \caption{Performance Evaluation of Different Embedding Sizes in terms of AUPRC Score }
  \label{fig:EmbSizeAUPRC}
\end{figure}

\begin{figure}[t!]
  \centering
  \includegraphics[width=1\linewidth]{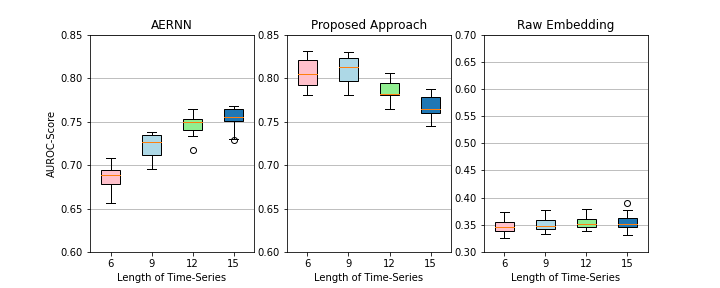}
  \caption{Performance Evaluation of Different Time-series Length in terms of AUROC Score }
  \label{fig:EmbSizeAUPRC}
\end{figure}

\begin{figure}[t!]
  \centering
  \includegraphics[width=1\linewidth]{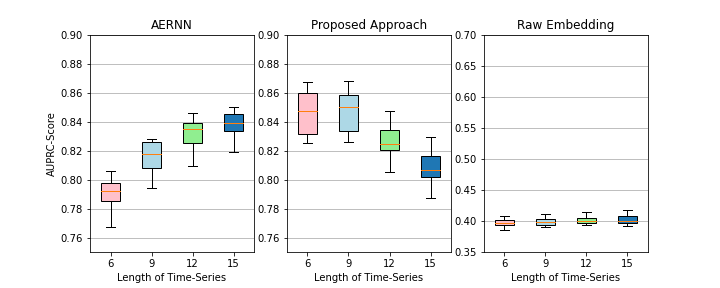}
  \caption{Performance Evaluation of Different Time-series Length in terms of AUPRC Score }
  \label{fig:EmbSizeAUPRC}
\end{figure}

We did experiments on $h \in \{2, 3, 4, 5, 6, 7, 8, 9\}$ to evaluate the effect of changing history length on the performance. To have the same condition for all experiments we set $dim = 128$ and $l = 12$. The result is shown in terms of the AUROC score in Fig. 6.

\begin{figure}[t!]
  \centering
  \includegraphics[width=1\linewidth]{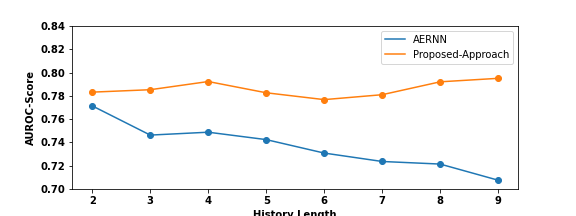}
  \caption{Performance Evaluation of Different History Lengths in terms of AUROC Score}
\end{figure}

The last series of experiments evaluated the proposed approach against recently proposed approaches for the problem of temporal link prediction in the literature. Table 2 demonstrates the results of applying multiple approaches to the Hyper dataset.

\begin{table}
\centering
\caption{COMPARISON OF PROPOSED APPROACH WITH STATE-OF-THE-ART (AUROC)}
\begin{tblr}{
  width = \linewidth,
  colspec = {Q[183]Q[119]Q[108]Q[119]Q[115]Q[283]},
  row{2} = {c},
  hline{2-3} = {-}{},
}
GraphSAGE & CTDNE  & SGNN   & TREND  & DMAB             & Proposed~Approach \\
0.5964    & 0.5742 & 0.5788 & 0.5948 & 0.6682\textbf{~} & \textbf{0.6843}   
\end{tblr}
\end{table}

\section{Discussion}
The first series of experiments have been designed in order to assess the performance of our proposed approach; these experiments confirm that our approach improves the standard temporal embedding model for co-authorship prediction by $17.34\%$ in terms of AUROC score. Also, results from Figs. $2$ and $3$ demonstrated how capturing velocity as an independent dynamic of nodes, rather than capturing dynamics altogether, has highly improved Raw Embedding quality. Figs. $2$ and $3$, represent for all $dim \in \{32, 64, 128, 256\}$, our approach has higher performance compared to the baseline. Also, we can see a gradual growth in the AUROC score of our approach by increasing the embedding size up to 128, and for the baseline, there is a sharp slope moving from 64 to 256. 
Fig. $2$ shows that increasing the embedding size improves the performance of the baseline while it doesn't make a notable change in the proposed approach's performance; while we get the best result for the baseline by setting the embedding size to 256, our approach has the best performance by setting the embedding size to 128; achieving higher performance with a smaller embedding size is more efficient in terms of time and computation. Fig. $3$ represents the results of link prediction for different embedding sizes in terms of AUPRC score for both approaches. We see that our approach has a better performance for all dimensions during all experiments on different samples. These series of experiments also reveal that our approach is highly robust to changes in embedding dimension as there is no more than 1\% change in performance. 

The next experiment was done to evaluate the effect of time-series length on learning improvement. We did experiments by setting $dim = 128$, and $h = 3$, and $l \in \{6, 9, 12, 15\}$. We can see in Fig.s $4$ and $5$ that including long-term dependencies in the learning phase has improved the prediction ability of AERNN as well as the Raw Embedding. This improvement is sharp for AERNN, moving from $l = 6$ to $l = 12$, while for our approach shorter time-series perform more efficiently. The improvement moving from the shortest series towards the longest one for AERNN is $6.66\%$ while this value for our approach is a negative value. This reveals that our approach captures enough predictive dependencies from shorter series by predicting velocity; so there is no need for having older histories of nodes in comparison to the baseline; learning from shorter series increases the efficiency of our approach and makes it a more general-purpose approach as in many datasets we do not have access to a long history of the network. 

The next experiment evaluated the effect of the recent history of node dynamics on its future behavior. As shown in Fig. 6, each method behaves in a different way in response to changing history length. The range of performance scores for different history values for our approach is [77.6, 79.5] while this range is [70.7, 77.1] for the baseline; these performance ranges demonstrate that our approach is mostly robust to the changes in history length and the difference between highest and lowest performance is 1.8 (this value for baseline is 6.3). Relying on longer history for better performance leads to a loss of generality and less efficient computations and in this case, the baseline is superior (although the proposed approach has better performance for all history lengths). The last experiment verified the efficiency of our approach against state-of-the-art for the task of temporal link prediction on a common temporal dataset.

Capturing high-order complex dependencies between nodes leads to high-performance link prediction. Appropriate structure, designed using neural networks, can capture these dependencies; in the meantime, there are some dependencies that we can capture more precisely and velocity is one of them. Based on experimental results, besides capturing the dynamics of the network generally, focusing on dynamics separately leads to capturing more predictive information; this way we improved embedding quality and got higher prediction performance. The other dynamic that we tried to directly take into account was \textit{local interaction of nodes with rest of the network}. We used the theory mentioned by Zhang et al. \cite{zhang2018link} to capture the hidden information in node interactions with the rest of the network. According to Zhang et al. \cite{zhang2018link}, we aggregated the location of the local neighborhood of each node to estimate its final location.

\subsection{Application}
Our proposed approach is not limited to co-authorship networks as we did not use co-authorship-specific features in the learning algorithm. The proposed method does not require special conditions, making it a general approach for time-series prediction problems. This approach provides enriched embedding and can outperform or improve other techniques in its class since it needs a short history length and a short series length. It can also be applied to current approaches as a final step in order to enrich currently generated embeddings. These upgraded embeddings can improve link prediction performance and be useful for other graph-based tasks; e.g., node classification. This method predicts a velocity curve, so more than $1$ step in the future may be predictable with high accuracy, making it superior to approaches in its class.

\section{Conclusion and Future Work}
This paper has proposed a novel method for the temporal link prediction problem with a special emphasis on co-authorship networks. Our approach leverages a sequence of author node embeddings and embedding dynamics to improve the prediction of prospective similarities between authors compared to the baseline: $0.1191$ absolute and $17.34\%$ relative improvement in the AUROC score. The novelty of our approach is in our formal accounting for embedding $velocity$ when making predictions; the velocity of author embeddings improves our ability to model prospective collaborations (i.e., links) as a function of the historical trajectory of author collaboration similarities. This method can, with minor modifications, be applied to other edge prediction tasks beyond co-authorship networks; also, it can be directly applied to embedding generated by various methods, as an improvement step.

Future directions for this work include the development of richer representations of authors including authors' research interests, similarities between scientific indexes, and institutions. Another direction of future research is trying to predict two or more steps in the future; as we predict the pattern of the velocity curve of the node's movement, we may be able to predict the state of the network for longer unseen steps in the future. To capture the dynamics of the movements of nodes in the embedding space more precisely, we want to take acceleration into account besides the velocity.

\section{Code Availability}
The source code of the proposed method required to reproduce the results is available at the public GitHub repository https://github.com/Sanaz11-3/Temporal-Link-Prediction.

\bibliographystyle{IEEEtran}

\end{document}